# Edge-aware Guidance Fusion Network for RGB–Thermal Scene Parsing

Wujie Zhou*, Shaohua Dong, Caie Xu, Yaguan Qian

School of Information and Electronic Engineering, Zhejiang University of Science & Technology, Hangzhou, China.
wujiezhou@163.com, shaohuadong2021@126.com, caiexu@163.com, qianyaguan@zust.edu.cn

## Abstract

RGB–thermal scene parsing has recently attracted increasing research interest in the field of computer vision. However, most existing methods fail to perform good boundary extraction for prediction maps and cannot fully use high-level features. In addition, these methods simply fuse the features from RGB and thermal modalities but are unable to obtain comprehensive fused features. To address these problems, we propose an edge-aware guidance fusion network (EGFNet) for RGB–thermal scene parsing. First, we introduce a prior edge map generated using the RGB and thermal images to capture detailed information in the prediction map and then embed the prior edge information in the feature maps. To effectively fuse the RGB and thermal information, we propose a multimodal fusion module that guarantees adequate cross-modal fusion. Considering the importance of high-level semantic information, we propose a global information module and a semantic information module to extract rich semantic information from the high-level features. For decoding, we use simple elementwise addition for cascaded feature fusion. Finally, to improve the parsing accuracy, we apply multitask deep supervision to the semantic and boundary maps. Extensive experiments were performed on benchmark datasets to demonstrate the effectiveness of the proposed EGFNet and its superior performance compared with state-of-the-art methods. The code and results can be found at https://github.com/ShaohuaDong2021/EGFNet.

## Introduction

Scene parsing is a fundamental technique in computer vision that aims to assign category labels to each of the pixels in a natural image. Hence, scene parsing has enhanced many applications in computer vision, such as autonomous driving (Wang et al. 2020; Zhou et al. 2021) and robot sensing (Zhang et al. 2018; Zhou et al. 2018; Zhou et al. 2021). In recent years, deep learning has become a promising solution to scene parsing. Existing methods based on fully convolutional networks have achieved noteworthy results (Lee et al. 2016; Chen et al. 2017; Luo et al. 2017; Zhang et al. 2017; Hou et al. 2018; Zhou et al. 2021). However, accurate scene parsing remains a challenge under poor light conditions. Some recent studies have noted this problem and proposed more robust methods via RGB–thermal scene parsing (Ha et al. 2017; Sun et al. 2019; Shivakumar et al. 2020; Sun et al. 2021; Zhang et al. 2021; Zhou et al. 2021). These methods use the complementary rich information and semantic information provided by the thermal images to RGB images under poor lighting conditions, thereby achieving high parsing performance.

Despite the abovementioned developments, some problems of RGB–thermal scene parsing remain to be solved. Owing to the lack of specific guidance on extracting boundaries, boundary preservation needs to be further improved. Existing methods based on fully convolutional networks reduce feature resolution, leading to loss of spatial details and distortion of object boundaries. In addition, existing methods use simple fusion strategies, such as elementwise addition or multiplication, thus failing to fully integrate multimodal information and undermining scene parsing performance. Moreover, most methods do not fully use high-level features with their rich semantic information. Therefore, a method to suitably extract and use high-level semantic information is desired.

To address these scene parsing problems, we propose an edge-aware guidance fusion network (EGFNet) for scene parsing based on an encoder–decoder architecture. The proposed EGFNet achieves remarkable performance for RGB–thermal scene parsing. We first introduce a method of embedding prior edge maps into the boundary features to enhance boundary information. To extract more information from the RGB and thermal features, we propose a multimodal fusion module (MFM) that integrates the multimodal features using efficient strategies. Unlike simple methods, such as fusion based on addition or concatenation, the MFM uses a complex fusion strategy to fully combine the information from the RGB and thermal modalities. In addition, a global information module (GIM) and a semantic information module (SIM) are proposed to extract high-level semantic information efficiently. Finally, we adopt multitask deep supervision to improve the segmentation performance. In general, the proposed EGFNet shows superior performance compared with state-of-the-art (SOTA) RGB–thermal scene parsing methods.

---

*Corresponding author.

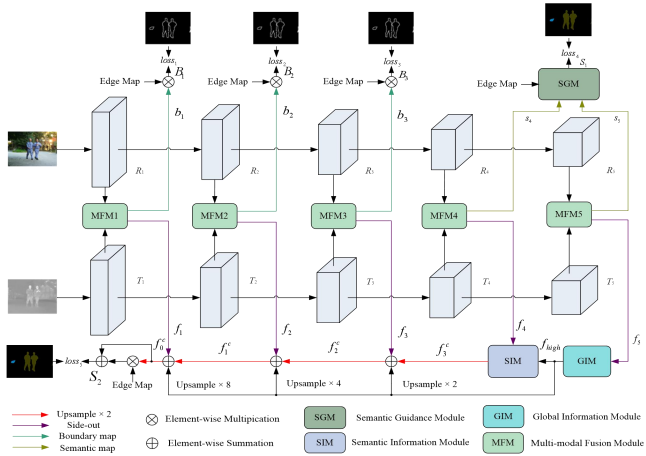

Figure 1: Architecture of proposed EGFNet for scene parsing.

The main contributions of the proposed EGFNet can thus be summarized as follows:

• The EGFNet is one of the pilot methods to use prior edge information for enhancing boundary extraction for RGB–thermal scene parsing and generating high-quality edge-aware prediction maps.

• We introduce the MFM to explore the effectiveness and complementarity between the RGB and thermal features. The MFM establishes a simple yet effective method to capture the complementarity of cross-modal features.

• To extract high-level semantic information, we propose the GIM and SIM, which fully and efficiently use high-level features.

• We adopt multitask deep supervision to obtain detailed object boundaries and improve parsing performances.

## Related work

In recent years, an increasing number of deep-learning-based scene parsing methods have been proposed and have achieved good performances. One of the essential aspects of these methods is the extraction of representative features. To this end, Yu et al. (2018) proposed a context path with fast downsampling to enlarge the receptive field. Pohlen et al. (2017) combined multiscale context with pixel-level accuracies using two processing streams within a neural network. Sun et al. (2019) proposed a network to maintain high-resolution representations throughout the parsing stages by connecting high-to-low-resolution convolutions in parallel. Yu et al. (2018) proposed a smooth network with a channel attention block to select the discriminative features. Romera et al. (2018) proposed a deep architecture using residual connections and factorized convolutions for efficient parsing with remarkable accuracy. Huang et al. (2019) proposed a criss-cross attention module to obtain rich contextual information. He et al. (2019) proposed a network that adaptively constructs multiscale contextual representations with multiple well-designed adaptive context modules.

Recently, single-modal methods, such as those mentioned above, have been improved by employing information from complementary modalities (e.g., depth maps and thermal images). Hazirbas et al. (2016) proposed a network to integrate multilevel depth features with an RGB encoder through a bottom-up approach to improve scene parsing. Wang et al. (2018) proposed depth-aware convolution and average pooling operations for RGB–depth scene parsing. Zhou et al. (2020) proposed a gate-fusion module to regularize feature fusion for detecting salient objects in RGB–depth images. Zhang et al. (2020) proposed a complementary interaction network to select useful representations from RGB images and their corresponding depth maps to integrate cross-modal features. Chen et al. (2020) proposed a disentangled cross-modal fusion module to extract structural and content representations from RGB images and depth maps. Wang et al. (2021) proposed a channelwise fusion module for multinetwork and multilevel selective fusion of RGB–depth parsing.

Boundary details can improve scene parsing substantially. To correct blurred boundaries, some methods extract specific boundary features. Zhang et al. (2020) proposed a boundary-guided deep neural network for scene parsing to suppress irrelevant boundary information while suitably localizing and exploring the structures of objects. Yang et al. (2021) devised edge feature enhancement to use edge-specific features efficiently. Wang et al. (2021) proposed a contour self-compensated module to generate accurate saliency maps with complete contours; the salient contours were then used as third labels for the ground truth. Kong et al. (2021) proposed an adversarial edge-aware image colorization approach combining multitask outputs with scene parsing. Unlike these methods, the proposed EGFNet uses a novel prior edge map to enhance the boundaries and improve scene parsing performance.

## Proposed EGFNet

### Architecture

The architecture of the proposed EGFNet is shown in Fig. 1. We use ResNet-152 (He et al. 2016) as the backbone of the encoders for both the RGB and thermal branches for extracting features. Owing to the high computational overhead, we use a $1 \times 1$ convolution to reduce the number of channels to 64. From low to high levels, the extracted RGB and thermal features are denoted as $R_i$ and $T_i$ ($i \in \{1, 2, ..., 5\}$), respectively.

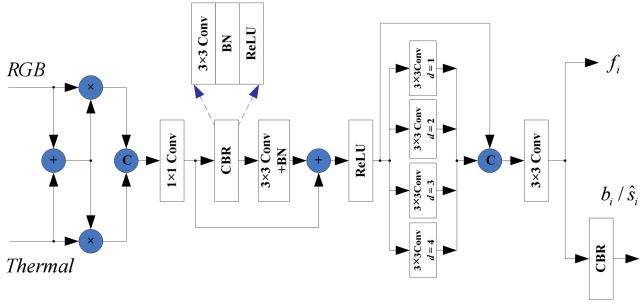

Figure 2: Architecture of proposed MFM.

In the encoder, we use the novel MFM to fuse complementary information from the RGB and thermal modalities. The MFM provides fusion features $f_i$ ($i \in \{1, 2, ..., 5\}$), boundary features $b_j$ ($j \in \{1, 2, 3\}$), and semantic features $s_t$ ($t \in \{4, 5\}$). We propose an edge-aware method to embed prior edge information in the feature maps and obtain clearer boundaries, thereby improving the parsing performance of EGFNet. In addition, the proposed GIM and SIM extract high-level semantic information.

In the decoder, we adopt a simple fusion module (SFM) to fuse cascaded features. By fusing the high-level semantic information and skip-connection features, we extract discriminative and comprehensive semantic information. Finally, we introduce multitask deep supervision for the semantic and boundary maps.

**Edge-aware guidance**

Edge-aware guidance for scene parsing aims to determine object boundaries in semantic maps accurately. Most existing methods use complex deep convolutional neural networks to capture the boundary features. To improve efficiency, we adopt a traditional edge-detection algorithm that allows obtaining details from the RGB and thermal images directly.

We first use the Sobel operator (Sobel et al. 1968) to extract the edge information from the RGB and thermal images. Then, we add the extracted edge information from the two modalities to fuse their distinct features and obtain a prior edge map. Finally, we embed the prior edge information in the boundary feature maps using elementwise multiplication to increase boundary accuracy in the prediction map. In addition, we improve the parsing performance by fusing the prior edge map and side-out semantic prediction map with the final semantic features, as shown in Fig. 1.

**Multimodal fusion module**

Exploring accurate multimodal fusion features is essential for achieving high-performance multimodal scene parsing. Thus, instead of using a simple fusion strategy, we propose the MFM to extract fusion features, thus outperforming simple feature concatenation or summation. The MFM architecture is shown in Fig. 2.

We first adopt elementwise summation of the features from the RGB and thermal modalities. Then, we apply various operations, including elementwise multiplication, channelwise concatenation, and convolution, to obtain complementary information as follows:

$$f_m = Conv_{1 \times 1}(Cat((R_i + T_i) \otimes R_i, (R_i + T_i) \otimes T_i)), i \in \{1,2,3,4,5\}, \quad (1)$$

where $Cat$ denotes concatenation, $\otimes$ denotes elementwise multiplication, $Conv_{1 \times 1}$ denotes $1 \times 1$ convolution, and $R_i$ and $T_i$ represent the side-out features of the RGB and thermal branches, respectively (Fig. 1).

We use residual learning to obtain deeper semantic features as follows:

$$\hat{f}_m = relu(f_m + BN(Conv_{3 \times 3}(CBR(f_m)))), i \in \{1,2,3,4,5\}. \quad (2)$$

First, the features pass through the convolution block $CBR$, and the $3 \times 3$ convolutions in $CBR$ are followed by the batch normalization and rectified linear unit (ReLU) layers, $BN$ and $relu$, respectively.

To enlarge the receptive field and extract a representative global context, we use atrous spatial pyramid pooling (Chen et al. 2018). Specifically, we construct four parallel dilated convolutions with rates $r = \{1, 2, 3, 4\}$ and combine the four sets of features with the input features using concatenation. Then, we use a $3 \times 3$ convolution to extract the fusion features $f_i$:

$$\begin{cases} \hat{f}_{m1} = Conv_{3 \times 3, rates=1}(\hat{f}_m) \\ \hat{f}_{m2} = Conv_{3 \times 3, rates=2}(\hat{f}_m) \\ \hat{f}_{m3} = Conv_{3 \times 3, rates=3}(\hat{f}_m) \\ \hat{f}_{m4} = Conv_{3 \times 3, rates=4}(\hat{f}_m) \end{cases} \quad (3)$$

$$f_i = Conv_{3 \times 3}(Cat(\hat{f}_m, \hat{f}_{m1}, \hat{f}_{m2}, \hat{f}_{m3}, \hat{f}_{m4})), i \in \{1,2,3,4,5\}. \quad (4)$$

Finally, except for the explicit usage of the semantic cues in EGFNet, the convolution block $CBR$ is applied to obtain detailed information and semantic information as follows:

$$b_i = CBR(f_i), i \in \{1,2,3\}, \quad (5)$$
$$s_i = CBR(f_i), i \in \{4,5\}. \quad (6)$$

**GIM and SIM**

Low-level features contain detailed information, and high-level features contain comprehensive semantic information (Zeiler et al. 2014). Accordingly, we first introduce the GIM (Fig. 3) and SIM (Fig. 4) to capture high-level semantic information and then fuse the cascaded multilevel cross-modal features using the SFM (Fig. 5).

The GIM is similar to atrous spatial pyramid pooling and aims to obtain discriminative semantic information as follows:

$$\hat{f}_{a0} = Conv_{1 \times 1}(f_5), \quad (7)$$

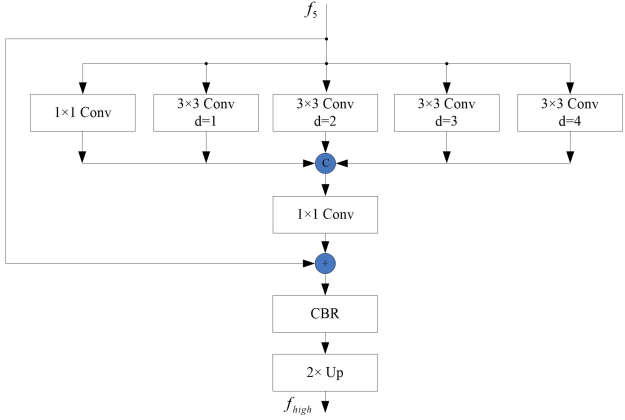

Figure 3: Architecture of proposed GIM.

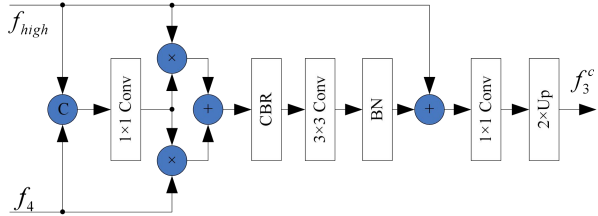

Figure 4: Architecture of proposed SIM.

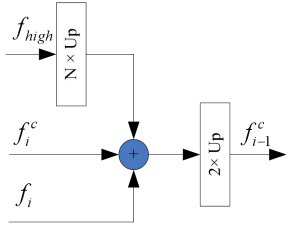

Figure 5: Architecture of proposed SFM.

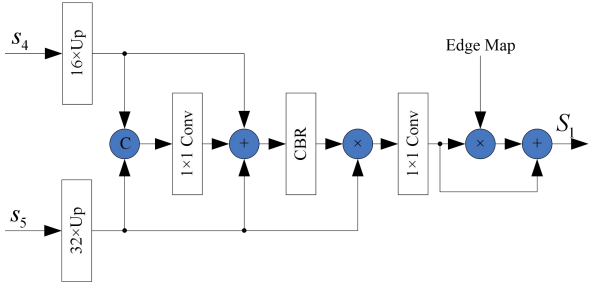

Figure 6: Architecture of proposed SGM.

$$\begin{cases} \hat{f}_{a1} = Conv_{3\times3, rates=1}(f_5) \\ \hat{f}_{a2} = Conv_{3\times3, rates=2}(f_5) \\ \hat{f}_{a3} = Conv_{3\times3, rates=3}(f_5) \\ \hat{f}_{a4} = Conv_{3\times3, rates=4}(f_5) \end{cases} \quad (8)$$

$$f_a = Conv_{1\times1}(Cat(\hat{f}_{a0}, \hat{f}_{a1}, \hat{f}_{a2}, \hat{f}_{a3}, \hat{f}_{a4})), \quad (9)$$

$$f_{high} = up_{\times2}(CBR(f_5 + f_a)), \quad (10)$$

where $up_{\times2}$ denotes upsampling by a factor of two.

Before applying the 1 × 1 convolutional layer to achieve comprehensive feature fusion, we combine $f_{high}$ and $f_4$ to learn complementary information. Moreover, we apply elementwise multiplication to eliminate redundant information. Finally, a residual connection is used to preserve the original information and generate the final features as follows:

$$f_{s1} = Conv_{1\times1}(Cat(f_{high}, f_4)), \quad (11)$$

$$f_{s2} = f_{s1} \otimes f_{high} + f_{s1} \otimes f_4, \quad (12)$$

$$f_i^c = up_{\times2}(Conv_{1\times1}(f_{high} + BN(Conv_{3\times3}(CBR(f_{s2}))))), i = 3. \quad (13)$$

We aggregate the multilevel features and high-level deep semantic information in a coarse-to-fine approach using the SFM to obtain the comprehensive features. We apply upsampling by a factor of $N$ ($up_{\times N}$) to $f_{high}$ such that it has the same size as $f_i$:

$$\hat{f}_{high} = up_{\times N}(f_{high}). \quad (14)$$

Finally, we fuse the features through elementwise summation:

$$f_{i-1}^c = up_{\times2}(\hat{f}_{high} + f_i^c + f_i), i \in \{1,2,3\}. \quad (15)$$

**Multitask deep supervision**

To obtain more accurate boundaries and distinct semantic features, we use multitask deep supervision to supervise the boundary and semantic maps.

We first resize $b_i$ to the same size as the edge map. Then, the prior edge map is embedded in the boundary feature map, as mentioned above. Hence, the boundary prediction map can be highlighted with a complete structure and sharp boundaries. This process can be formulated as follows (Fig. 1):

$$B_i = up_{\times N}(b_i) \otimes edge, i \in \{1,2,3\}. \quad (16)$$

Moreover, we apply the prior edge map to the intermediate and final semantic prediction maps. To improve learning from side-out semantic information, we propose a semantic guidance module (SGM) that fuses the corresponding features efficiently.

The SGM architecture is detailed in Fig. 6. We first upsample semantic maps ŝ$_1$ and ŝ$_2$ such that they have the same size as the ground-truth map. Then, we fuse the features through concatenation followed by a 1 × 1 convolution:

$$\hat{f}_{sem1} = Conv_{1\times1}(Cat(up_{\times16}(s_4), up_{\times32}(s_5))), \quad (17)$$

where $up_{\times16}$ and $up_{\times32}$ denote upsampling by factors of 16 and 32 using bilinear interpolation, respectively. Then, the side-out semantic prediction is generated as follows:

$$\hat{f}_{sem2} = \hat{f}_{sem1} + up_{\times16}(s_4) + up_{\times32}(s_5), \quad (18)$$

$$\hat{f}_{sem} = Conv_{1\times1}(CBR(\hat{f}_{sem2}) \otimes up_{\times32}(s_5)). \quad (19)$$

We then embed the prior edge information in the side-out semantic prediction as follows:

Table 1. Results on the MFNet dataset. Each value in boldface indicates the best result for the corresponding column.

| Methods | Car | | Person | | Bike | | Curve | | Car Stop | | Guardrail | | Color Cone | | Bump | | mAcc | mIou |
|---|---|---|---|---|---|---|---|---|---|---|---|---|---|---|---|---|---|---|
| | Acc | Iou | Acc | Iou | Acc | Iou | Acc | Iou | Acc | Iou | Acc | Iou | Acc | Iou | Acc | Iou | | |
| FRRN | 80.0 | 71.2 | 53.0 | 46.1 | 65.1 | 53.0 | 34.0 | 27.1 | 21.6 | 19.1 | 0.0 | 0.0 | 34.7 | 32.5 | 36.2 | 30.5 | 47.1 | 41.8 |
| BISeNet | 90.0 | 84.5 | 65.0 | 54.3 | 75.0 | 61.4 | 32.1 | 25.7 | 32.3 | 26.2 | 3.2 | 0.9 | 49.6 | 43.3 | 48.1 | 40.5 | 54.9 | 48.2 |
| DFN | 90.7 | 81.4 | 67.7 | 52.8 | 71.5 | 57.5 | 49.2 | 34.9 | 35.1 | 23.8 | 4.1 | 0.9 | 44.2 | 31.0 | 54.6 | 47.5 | 57.3 | 47.5 |
| SegHRNet | 92.2 | 86.6 | 73.1 | 59.8 | 74.9 | 61.3 | 47.0 | 33.2 | 38.3 | 28.7 | 7.3 | 1.4 | 54.6 | 47.2 | 61.5 | 46.2 | 60.9 | 51.3 |
| CCNet | 86.7 | 79.5 | 59.4 | 52.7 | 66.0 | 56.2 | 39.2 | 32.2 | 34.8 | 29.0 | 1.3 | 1.2 | 45.7 | 41.0 | 0.2 | 0.2 | 48.1 | 43.3 |
| APCNet | 89.8 | 83.0 | 61.3 | 51.6 | 73.4 | 58.7 | 37.1 | 27.0 | 35.6 | 30.3 | 36.1 | **11.8** | 41.4 | 35.6 | 50.7 | 45.6 | 58.3 | 49.0 |
| MFNet | 77.2 | 65.9 | 67.0 | 58.9 | 53.9 | 42.9 | 36.2 | 29.9 | 12.5 | 9.9 | 0.1 | 0.0 | 30.3 | 25.2 | 30.0 | 27.7 | 45.1 | 39.7 |
| FuseNet | 81.0 | 75.6 | 75.2 | 66.3 | 64.5 | 51.9 | 51.0 | 37.8 | 17.4 | 15.0 | 0.0 | 0.0 | 31.1 | 21.4 | 51.9 | 45.0 | 52.4 | 45.6 |
| DepthAwareCNN | 85.2 | 77.0 | 61.7 | 53.4 | 76.0 | 56.5 | 40.2 | 30.9 | 41.3 | 29.3 | 22.8 | 8.5 | 32.9 | 30.1 | 36.5 | 32.3 | 55.1 | 46.1 |
| RTFNet | 93.0 | 87.4 | 79.3 | 70.3 | 76.8 | 62.7 | 60.7 | **45.3** | 38.5 | 29.8 | 0.0 | 0.0 | 45.5 | 29.1 | **74.7** | **55.7** | 63.1 | 53.2 |
| FuseSeg-161 | 93.1 | **87.9** | 81.4 | **71.7** | 78.5 | **64.6** | 68.4 | 44.8 | 29.1 | 22.7 | **63.7** | 6.4 | 55.8 | 46.9 | 66.4 | 47.9 | 70.6 | 54.5 |
| ABMDRNet | 94.3 | 84.8 | **90.0** | 69.6 | 75.7 | 60.3 | 64.0 | 45.1 | 44.1 | 33.1 | 31.0 | 5.1 | 61.7 | 47.4 | 66.2 | 50.0 | 69.5 | 54.8 |
| Ours | **95.8** | 87.6 | 89.0 | 69.8 | **80.6** | 58.8 | **71.5** | 42.8 | **48.7** | **33.8** | 33.6 | 7.0 | **65.3** | **48.3** | 71.1 | 47.1 | **72.7** | **54.8** |

$$S_1 = edge \otimes \hat{f}_{sem} + \hat{f}_{sem}. \quad (20)$$

Similarly, we enhance the final semantic prediction of EGFNet as follows:

$$S_2 = edge \otimes f_0^c + f_0^c. \quad (21)$$

Boundary maps $B_1$, $B_2$, and $B_3$ and semantic maps $S_1$ and $S_2$ are supervised by the ground truth using the weighted cross-entropy loss with weights set as in the study by Paszke et al. (2016):

$$L = -\frac{1}{W \times H} \sum_{x=1}^{W} \sum_{y=1}^{H} W_{eight}\left(T_{x,y} \log(P_{x,y}) + (1-T_{x,y})\log(1-P_{x,y})\right), \quad (22)$$

where $W$ and $H$ are the width and height of the image, respectively, and $T$ and $P$ denote the ground-truth and prediction maps, respectively. The variable $W_{eight}$ denotes the boundary weight while calculating the boundary loss and describes the semantic weight when calculating the semantic loss.

The total loss for multitask deep supervision is defined as

$$L_{total} = \sum_{i=1}^{3} L_i + \sum_{j=4}^{5} L_j, \quad (23)$$

where $L_i$ is the boundary loss and $L_j$ is the semantic loss (Fig. 1).

## Experimental results

We evaluated the proposed EGFNet and compared it with SOTA scene parsing methods through extensive experiments on two public datasets. We also conducted ablation studies to demonstrate the effectiveness of the various components of EGFNet.

**Datasets**

We trained the EGFNet on the MFNet (Ha et al. 2017) and PST900 (Shivakumar et al. 2020) datasets. The MFNet dataset contains 1569 pairs of RGB and thermal images, with 820 pairs corresponding to daytime scenes and 749 pairs corresponding to nighttime scenes. The dataset comprises nine classes, including the background. The resolution of the image pairs is 480 × 640 pixels. We followed the training, testing, verification, and dataset splitting approaches used by Ha et al. (2017). The PST900 dataset contains 894 aligned pairs of RGB and thermal images with pixel-level human annotations comprising five semantic classes, including the background. We used the splitting approach proposed by Shivakumar et al. (2020) and resized each input image to 640 × 1280 pixels.

**Training details**

We used the PyTorch 1.7.0, CUDA 10.0, and cuDNN 7.6 libraries to implement the proposed EGFNet. A computer equipped with an Intel 3.6 GHz i7 CPU and a single NVIDIA TITAN Xp graphics card was used for training and testing. As the graphics card memory was limited to 12 GB, we adjusted the batch sizes for different evaluated networks accordingly.

For training, we used data augmentation operations such as random flipping and cropping. The parameters of the backbone were initialized based on the ResNet-152 model (He et al. 2016). We trained EGFNet for 400 epochs and used the Ranger optimizer with an initial learning rate and weight decay of 5e−5 and 5e−4, respectively. We also used the weighted cross-entropy for both the semantic and boundary loss functions as well as weighting detailed by

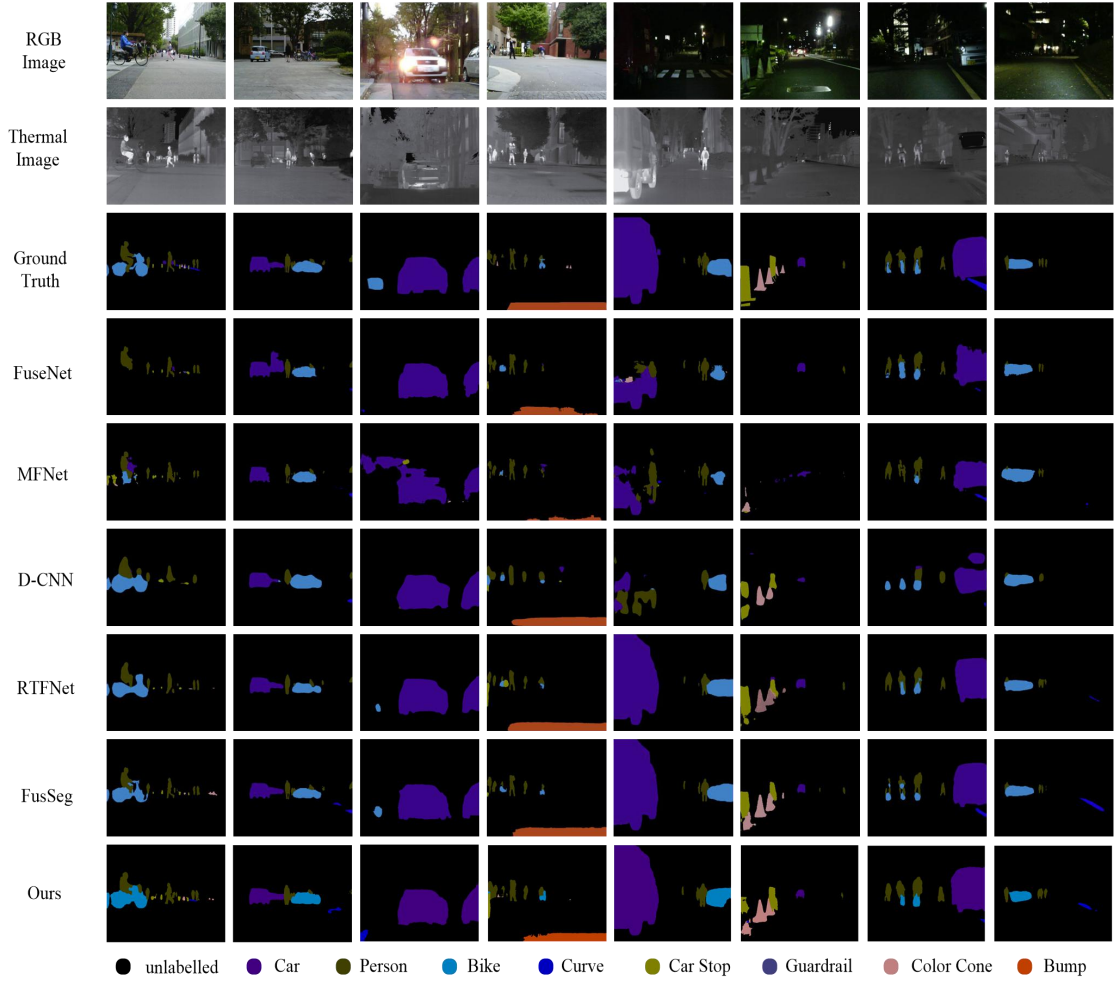

Figure 7: Segmentation results of fusion modules in typical nighttime and daytime RGB-T images shown in the right four and left four columns, respectively. The proposed EGFNet provides better segmentation under varying lighting conditions than the comparison networks.

Paszke et al. (2016).

**Evaluation metrics**

For the quantitative evaluations, we adopted some widely used evaluation metrics, including mean intersection over union (mIoU) and mean accuracy (mAcc), to evaluate the performances of different scene parsing methods.

**Comparative results**

For the MFNet dataset (Ha et al. 2017), we compared the proposed EGFNet with FRRN (Pohlen et al. 2017), BiSeNet (Yu et al. 2018), DFN (Yu et al. 2018), SegHRNet (Ke et al. 2019), MFNet (Ha et al. 2017), FuseNet (Hazirbas et al. 2016), DepthAwareCNN (Wang et al. 2018), RTFNet (Sun et al. 2019), FuseSeg-161 (Sun et al. 2021), APCNet (He et al. 2019), CCNet (Huang et al. 2019), and ABMDRNet (Zhang et al. 2021). The quantitative results are summarized in Table 1 and demonstrate that our method outperforms other SOTA methods on the MFNet dataset. To further evaluate the proposed network, we tested it with the daytime and nighttime RGB-T images; Table 2 summarizes the comparative results.

The visual comparison results are collated in Fig. 7, and we observe that our network provides superior results under various challenging lighting conditions compared with other SOTA methods for the MFNet dataset.

We designed additional experiments to prove the effectiveness of the proposed network on the PST900 dataset (Shivakumar et al. 2020). We compared the results from our network with those of CCNet (Huang et al. 2019), ACNet (Hu et al. 2019), EFFicient FCN (Liu et al. 2020), RTFNet (Sun et al. 2019), and PSTNet (Shivakumar et al. 2020). The results summarized in Table 3 indicate the excellent applicability of the proposed approach.

Table 3. Results on the PST900 dataset.

| Methods | Background | | Hand-Drill | | Backpack | | Fire-Extinguisher | | Survivor | | mAcc | mIoU |
|---|---|---|---|---|---|---|---|---|---|---|---|---|
| | Acc | IoU | Acc | IoU | Acc | IoU | Acc | IoU | Acc | IoU | | |
| Efficient FCN | 99.81 | 98.63 | 32.08 | 30.12 | 60.06 | 58.15 | 78.87 | 39.96 | 32.76 | 28.00 | 60.72 | 50.98 |
| CCNet | **99.86** | 99.05 | 51.77 | 32.27 | 68.30 | 66.42 | 67.79 | 51.84 | 60.84 | 57.50 | 69.71 | 61.42 |
| ACNet | 99.83 | 99.25 | 53.59 | 51.46 | 85.56 | **83.19** | 84.88 | 59.95 | 69.10 | 65.19 | 78.67 | 71.81 |
| RTFNet | 99.78 | 99.02 | 7.79 | 7.07 | 79.96 | 74.17 | 62.39 | 51.93 | 78.51 | 70.11 | 65.69 | 60.46 |
| PSTNet | - | 98.85 | - | 53.60 | - | 69.20 | - | 70.12 | - | 50.03 | - | 68.36 |
| Ours | 99.48 | **99.26** | **97.99** | **64.67** | **94.17** | 83.05 | **95.17** | **71.29** | **83.30** | **74.30** | **94.02** | **78.51** |

## Ablation study

To demonstrate the effectiveness of the key components of the proposed EGFNet, we applied the same network parameters for retraining each ablation experiment on the MFNet dataset, and these results are listed in Table 4.

**Effectiveness of prior edge information:** To demonstrate the effectiveness of edge information, we designed a variant without implicitly using edge cues in EGFNet, denoting it as w/o edge. The corresponding results are listed in Table 4. The variant w/o edge exhibits worse performance compared to the EGFNet with edge information, demonstrating the importance of edge information for scene parsing and validating its use.

**Effectiveness of MFM:** To demonstrate the

Table 2. Results from nighttime and daytime images.

| Methods | Daytime | | Nighttime | |
|---|---|---|---|---|
| | mAcc | mIoU | mAcc | mIoU |
| FRRN | 45.1 | 40.0 | 41.6 | 37.3 |
| BiSeNet | 52.1 | 44.5 | 50.3 | 45.0 |
| DFN | 53.7 | 42.2 | 52.4 | 44.6 |
| SegHRNet | 59.7 | 47.2 | 55.7 | 49.1 |
| CCNet | 55.3 | 43.5 | 42.4 | 38.1 |
| APCNet | 55.4 | 42.4 | 54.7 | 46.4 |
| MFNet | 42.6 | 36.1 | 41.4 | 36.8 |
| FuseNet | 49.5 | 41.0 | 48.9 | 43.9 |

Table 4. Results of ablation experiments.

| | mAcc | mIoU |
|---|---|---|
| Model (w/o edge) | 68.9 | 54.1 |
| Model (w/o MFM) | 68.1 | 53.1 |
| Model (w/o GIM) | 71.8 | 53.5 |
| Model (w/o SIM) | 69.1 | 53.2 |
| Model (w/o GIM & SIM) | 71.4 | 54.0 |
| Model (w/o SUP) | 71.7 | 53.3 |
| Model (Ours) | **72.7** | **54.8** |

effectiveness of the MFM, we replaced it with simple addition and denoted this variant as w/o MFM. As summarized in Table 4, the proposed EGFNet performs better than the variant w/o MFM, demonstrating the reliability of the module.

**Effectiveness of GIM and SIM:** To demonstrate the effectiveness of GIM and SIM, we designed three ablation experiments by removing GIM (denoted as w/o GIM), removing SIM (denoted as w/o SIM), and removing both GIM and SIM (denoted as w/o GIM & SIM). We applied simple addition when removing each module. The three evaluated variants provided declined performances than when using the SIM and GIM in EGFNet. These results indicate the importance of the GIM and SIM for obtaining high-level semantic information.

**Effectiveness of multitask deep supervision:** To demonstrate the efficiency of multitask deep supervision, we removed all the supervision except for the final supervision stage while maintaining all the other network parameters (denoted as w/o SUP). Table 4 indicates that the EGFNet performance considerably decreases when only one supervision stage is used.

## Conclusion

We propose the EGFNet for RGB–thermal scene parsing. We demonstrate that prior edge information contributes toward generating high-quality and comprehensive scene parsing maps. Moreover, the MFM enables exploitation of the complementarity between the RGB and thermal modalities, while the GIM and SIM allow extraction of high-level semantic information. Furthermore, the proposed multitask deep supervision promotes effective and robust scene parsing. Experiments were performed with two benchmark datasets to demonstrate the high performance of the EGFNet, and the results from ablation experiments verify the contributions of the most important network components.

## Acknowledgment

This work was supported by the National Natural Science Foundation of China (Grant No. 61502429).